\documentclass[letterpaper]{article} 
\usepackage{aaai24}  
\usepackage{times}  
\usepackage{helvet}  
\usepackage{courier}  
\usepackage[hyphens]{url}  
\usepackage{graphicx} 
\usepackage{natbib}  
\usepackage{caption} 
\usepackage{algorithm}
\usepackage{algorithmic}

\usepackage{newfloat}
\usepackage{listings}
\DeclareCaptionStyle{ruled}{labelfont=normalfont,labelsep=colon,strut=off} 
\floatstyle{ruled}
\newfloat{listing}{tb}{lst}{}
\floatname{listing}{Listing}
\usepackage{algorithm}
\usepackage{algorithmic}

 \usepackage{multirow}
 \usepackage{amsfonts,amssymb}
 \usepackage{amsmath}
\title{High-Order Structure Based Middle-Feature Learning for \\Visible-Infrared Person Re-Identification}
\author {
    Liuxiang Qiu\textsuperscript{\rm 1,2}\equalcontrib,
    Si Chen\textsuperscript{\rm 2}\equalcontrib,
    Yan Yan\textsuperscript{\rm 1}\thanks{Corresponding author (email: yanyan@xmu.edu.cn).},
    Jing-Hao Xue\textsuperscript{\rm 3},
    Da-Han Wang\textsuperscript{\rm 2},
    Shunzhi Zhu\textsuperscript{\rm 2}
}
\affiliations {
    \textsuperscript{\rm 1}Xiamen University, China ~
    \textsuperscript{\rm 2} Xiamen University of Technology, China ~
    \textsuperscript{\rm 3} University College London, UK
}

\usepackage{bibentry}

\begin{document}

\maketitle

\begin{abstract}
Visible-infrared person re-identification (VI-ReID) aims to retrieve images of the same persons captured by visible (VIS)  and infrared (IR) cameras. 
Existing VI-ReID methods ignore high-order structure information of features while being relatively difficult to learn a reasonable common feature space due to the large modality discrepancy between VIS and IR images.  To address the above problems, we propose a novel high-order structure based middle-feature learning
network (HOS-Net) for effective VI-ReID. Specifically, we first leverage a short- and long-range feature extraction (SLE) module to effectively exploit both 
short-range and long-range features. Then, we propose a high-order structure learning (HSL) module to successfully 
 model the high-order relationship across different local features of each person image based on a whitened hypergraph network.
 This greatly alleviates model collapse and enhances feature representations. Finally, we develop a common feature space learning (CFL) module to learn a discriminative and reasonable common feature space based on middle features generated by aligning features from different modalities and ranges. In particular, a modality-range identity-center contrastive (MRIC) loss is proposed to reduce the distances between the VIS, IR, and middle features, smoothing the training process. Extensive experiments on the SYSU-MM01, RegDB, and LLCM  datasets show that our  HOS-Net achieves superior state-of-the-art performance. Our code is available at
\url{https://github.com/Jaulaucoeng/HOS-Net}.
\end{abstract}

\section{Introduction}

Over the past few years, person re-identification (ReID) has attracted increasing attention due to its significant importance in surveillance
and security applications. A large number of single-modality person ReID methods have been proposed based on visible (VIS) cameras. However, these methods may fail under low-light conditions. Unlike VIS cameras, infrared (IR) cameras are less affected by illumination changes. Recently, visible-infrared person re-identification (VI-ReID), which leverages both VIS and IR cameras, has been developed to match cross-modality person images, mitigating the limitations of single-modality person ReID.

A major challenge of VI-ReID is the large modality discrepancy between VIS and IR images. To reduce the modality discrepancy, existing VI-ReID methods can be divided into image-level and feature-level methods.  The image-level methods \cite{dai2018cross, wang2020cross, wei2022rbdf},  generate middle-modality or new modality images based on generative adversarial networks (GAN). However, the GAN-based methods easily suffer from the problems of color inconsistency or loss of image details. Hence, 
the generated images may not be reliable for subsequent classification.

The feature-level methods \cite{ye2021deep,lu2022learning,zhang2022dual} follow a two-step learning pipeline (i.e.,
they first extract features for VIS and IR images, and then 
map these features into a common feature space). Generally, these methods have two issues. 
On the one hand, they often ignore high-order structure information of features (i.e., the different levels of dependence across  local features), which can be important for matching cross-modality images. 
On the other hand, they usually  directly minimize the distances between VIS and IR features in the common feature space. However,  such a manner  increases the difficulty of learning a reasonable common feature space due to the large modality discrepancy.

To address the above issues, in this paper, we propose a novel high-order structure based {middle-feature} learning network (HOS-Net), which consists of a backbone, a short- and long-range feature extraction (SLE) module, a high-order structure learning (HSL) module, and a common feature space learning (CFL) module, for VI-ReID.  The key novelty of our method lies in the novel formulation of exploiting \textit{high-order structure information} and \textit{middle features} to learn a discriminative and reasonable common feature space, greatly alleviating the modality discrepancy.

Specifically, given a VIS-IR image pair, the SLE module (consisting of a convolutional branch and a Transformer branch) extracts short-range and long-range features. 
Then, the HSL module 
models the dependence on short-range and long-range features based on a whitened hypergraph. 
Finally, the CFL module learns a common feature space by generating and leveraging middle features. In the CFL module, instead of directly adding or concatenating features from different modalities and ranges, we leverage graph attention to properly align these features, obtaining middle features. 
Based on it, a modality-range identity-center contrastive (MRIC) loss is developed to reduce the distances between the VIS, IR, and middle features, smoothing the process of learning the common feature space.

The contributions of our work are twofold: 
\begin{itemize}

    \item First, we introduce an HSL module to learn high-order structure information of both short-range and long-range features. Such an innovative way effectively models high-order relationship across different local features of a person image without suffering from model collapse,  greatly enhancing feature representations.

    \item Second, we design a CFL module to learn a discriminative and reasonable common feature space by taking advantage of middle features. In particular,  a novel MRIC loss is developed to minimize the distances between VIS, IR, and middle features. This is beneficial for the extraction of  discriminative modality-irrelevant ReID features.
\end{itemize}

   Extensive experiments on the SYSU-MM01, RegDB, and LLCM datasets demonstrate that our proposed HOS-Net obtains excellent performance in comparison with several state-of-the-art VI-ReID methods.

\section{Related Work}
\noindent \textbf{Single-Modality Person Re-Identification (ReID).}
A variety of single-modality person ReID methods have been developed and achieved promising performance in the cases of occlusion, cloth-changing, and pose changes. Yan \textit{et al.} \cite{yan2021occluded} propose an occlusion-based {data augmentation} strategy and a bounded exponential distance loss for occluded person ReID. 
Jin \textit{et al.} \cite{jin2022cloth} introduce an additional gait recognition task to learn cloth-agnostic features. Note that these methods are based on VIS cameras and thus they perform poorly in low-light conditions.

\noindent \textbf{Visible-Infrared Person  Re-Identification (VI-ReID).} 
The image-level methods \cite{dai2018cross, wang2020cross, wei2022rbdf} often reduce the modality discrepancy by generating middle-modality images or new modality images. Wei \textit{et al.} \cite{wei2022rbdf} propose a bidirectional image translation subnetwork to generate middle-modality images from VIS and IR modalities.  
Li \textit{et al.} \cite{li2020infrared} and Zhang \textit{et al.} \cite{zhang2021towards} introduce light-weight middle-modality image generators to mitigate the modality discrepancy.  
Instead of generating middle-modality images, we align the features from different modalities and ranges with graph attention, generating reliable middle features. Moreover, we design an MRIC loss to optimize the distances between VIS, IR, and middle features, benefiting the extraction of discriminative ReID features.

The feature-level methods map the features of different modalities into a common feature space to reduce the modality discrepancy. A few methods \cite{ye2021deep,yang2023top,lu2022learning} leverage  CNN or ViT as the backbone to extract features.
Some methods \cite{chen2022structure,wan2023g2da} adopt off-the-shelf key point extractors to generate key point labels of person images and learn modality-irrelevant features. But the key point extractor may introduce noisy labels, deteriorating the discriminability of final ReID features. Many VI-ReID methods \cite{liu2020parameter, huang2022cross, huang2023exploring} employ the contrastive-based loss, which directly minimizes the distances between VIS and IR features, to obtain a common feature space. However, 
it is not a trivial task to learn a reasonable common feature space due to the large modality discrepancy between modalities.

Our method belongs to feature-level methods. 
However, conventional feature-level methods mainly consider first-order structure information of features (i.e., the pairwise relation across  features). Moreover, they directly reduce the distances between VIS and IR features.
Different from these methods, our method not only captures high-order structure information of features but also generates middle features, greatly facilitating our model to learn an effective common feature space.

    \begin{figure*}[t]
    \centering
    \includegraphics[width=0.9 \linewidth]{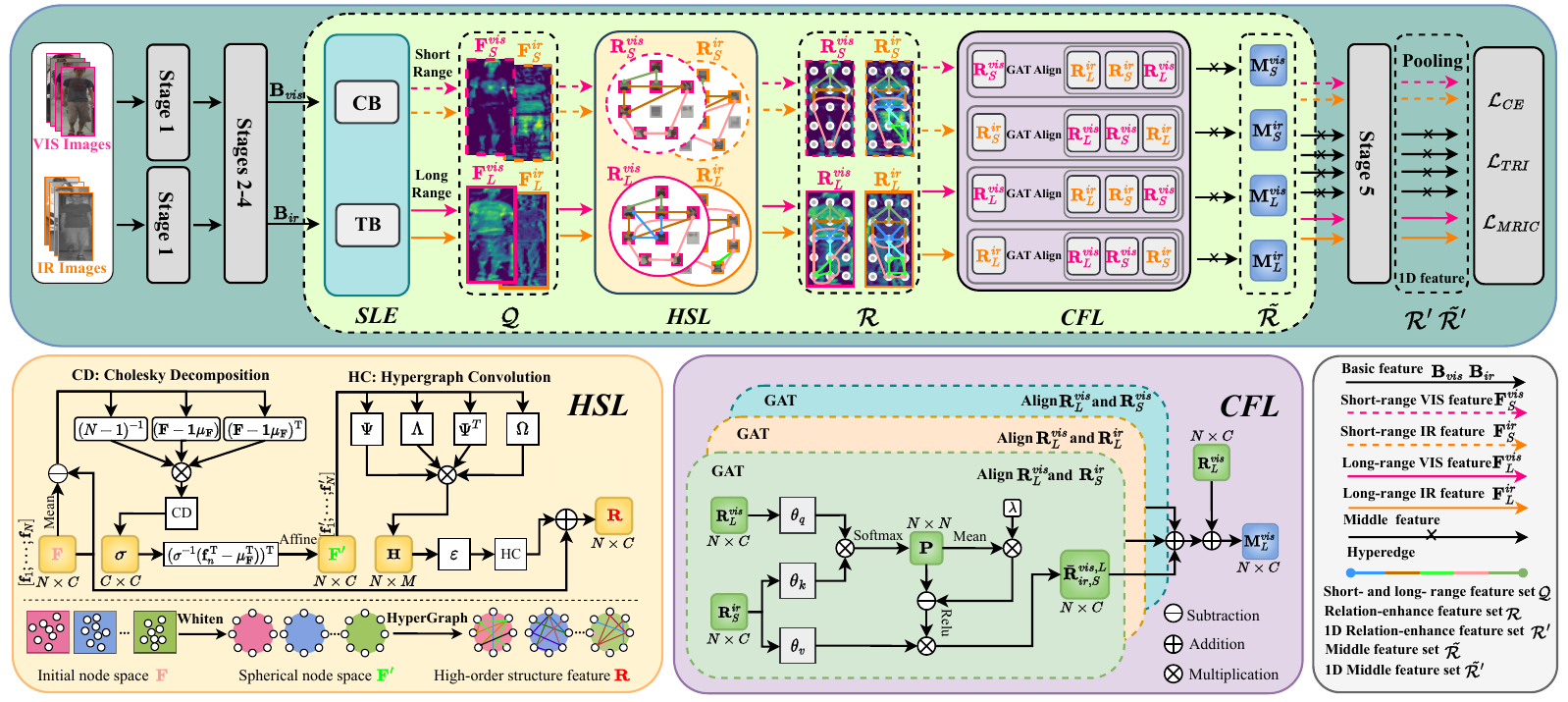}
    \caption{Overview of the proposed HOS-Net, including a backbone, a short- and long-range feature extraction (SLE) module, a high-order structure learning (HSL) module, and a common feature space learning (CFL) module. The HOS-Net is jointly optimized by $\mathcal{L}_{CE}$, $\mathcal{L}_{TRI}$, and $\mathcal{L}_{MRIC}$.}
    \label{F2}
    \end{figure*}

\noindent \textbf{Graph Neural Networks in Person Re-Identification.}
Graph neural network (GNN) is a class of neural networks that is designed to operate on graph-structured data.  
Li \textit{et al.} \cite{li2021decoupled} propose a pose and similarity based-GNN to reduce the problem of pose misalignment for single-modality person ReID. Wan \textit{et al.} \cite{wan2023g2da} develop a geometry guided dual-alignment strategy to align VIS and IR features, improving the consistency of multi-modality node representations. 
Different from pairwise connections in the vanilla graph models, Feng \textit{et al.} \cite{feng2019hypergraph} propose a hypergraph neural network (HGNN) to encode high-order feature correlations in a hypergraph structure.  Lu \textit{et al.} \cite{lu2023exploring}  model high-order spatio-temporal correlations  based on HGNN (which relies on high-quality skeleton labels) for video person ReID. 

{However, the above} HGNN-based methods may easily suffer from the model collapse problem (i.e., high-order correlations collapse to a single correlation) since a hyperedge can connect an arbitrary number of nodes. Unlike the above methods, 
we leverage the whitening operation, which plays the role of ``scattering'' on the nodes of the hypergraph, to significantly alleviate model collapse.

\section{Proposed Method}
\subsection{Overview}
The overview of our proposed HOS-Net is given in Figure \ref{F2}.
HOS-Net consists of a backbone, an SLE module, an
HSL module, and a CFL module.
In this paper, we adopt 
a two-stream AGW \cite{ye2021deep} as the backbone. Given a  VIS-IR image pair with the same identity, we first pass it through the backbone to obtain paired VIS-IR features.
Then, these features are fed into the SLE module to learn short-range and long-range features for each modality. Next, the HSL module exploits high-order structure information of short-range and long-range features based on a whitened hypergraph network. Finally, {the CFL module learns a discriminative} common feature space based on middle features that are obtained by aligning VIS and IR features through graph attention. In the CFL module, {we develop an} MRIC loss  to reduce the distances between the VIS, IR, and middle features, greatly smoothing the process of learning the common feature space.
\subsection{Short- and Long-Range Feature Extraction (SLE) Module}
Conventional VI-ReID methods \cite{ye2021deep, yang2023top} often leverage CNN or  ViT for feature extraction. CNN excels at capturing short-range features, while ViT is good at obtaining long-range features \cite{zhang2022parc, chen2022mobile}.
In this paper, we adopt an SLE module to exploit short-range and long-range features by taking advantage of both CNN and ViT. The SLE module contains a convolutional branch (CB) and a Transformer branch (TB).  
CB contains 3  convolutional blocks and TB contains 2 Transformer blocks with 4 heads. 
Assume that we have a VIS-IR image pair $\{\mathbf{I}^{vis}, \mathbf{I}^{ir} \}$ with the same identity. We denote the VIS and IR features obtained  from the backbone as 
 $\mathbf{B}^{vis}$ and $\mathbf{B}^{ir}$, respectively.

Then, $\mathbf{B}^{vis}$ and $\mathbf{B}^{ir}$ are fed into the SLE module to obtain short-range and long-range features for each modality, i.e., 
    \begin{equation} \label{E1}
    \begin{split} 
        &\  \mathbf{F}^{vis}_{S}    =\mathrm{CB}(\mathbf{B}^{vis}),     \quad  \mathbf{F}^{vis}_{L} =\mathrm{TB}(\mathbf{B}^{vis}),  \\
         &\  \mathbf{F}^{ir}_{S}  \, =\mathrm{CB}(\mathbf{B}^{ir}),    \quad  
 \,  \mathbf{F}^{ir}_{L} \, =\mathrm{TB}(\mathbf{B}^{ir}),  \\
    \end{split}
    \end{equation}
where  $\mathrm{CB}(\cdot)$ and $\mathrm{TB}(\cdot)$ represent the convolutional branch and the Transformer branch, respectively; $\mathbf{F}^{vis}_{S} \in \mathbb{R}^{H\times W\times C}$ and $\mathbf{F}^{ir}_{S} \in \mathbb{R}^{H\times W\times C}$ denote the short-range features for the VIS and IR images, respectively;  $\mathbf{F}^{vis}_{L} \in \mathbb{R}^{H\times W\times C}$ and $\mathbf{F}^{ir}_{L} \in \mathbb{R}^{H\times W\times C}$ denote the long-range features for the VIS and IR images, respectively; $H$, $W$, and $C$ denote the height, width, and channel number of the feature, respectively. 
Thus, for a VIS-IR image pair, we can obtain the feature set $\mathcal{Q}=\{ 
\mathbf{F}^{vis}_{L}, \mathbf{F}^{vis}_{S}, \mathbf{F}^{ir}_{L}, \mathbf{F}^{ir}_{S} \}$, which is used as the input of the HSL module.

\subsection{High-Order Structure Learning (HSL) Module}
The features extracted from the SLE module only encode pixel-wise and region-wise dependencies in the person images. However, the high-order structure information, which indicates different levels of relation in the features (e.g., head, torso, upper
arm, and lower arm belongs to the upper part of the body while head, torso, arm, and leg belong to the whole body), is not well exploited. Therefore, inspired by HGNN \cite{feng2019hypergraph}, we introduce the HSL module to capture high-order correlations across different local features, enhancing feature representations. Note that the conventional HGNN tends to suffer from the problem of model collapse. To alleviate this problem, we take advantage of the whitening operation and apply it to the hypergraph network, as shown in Figure  \ref{F2}.

Different from pairwise connections in the vanilla graph models, a hypergraph can connect an arbitrary number of nodes to exploit high-order structure information.   We construct a whitened hypergraph for each feature in $\mathcal{Q}$. 
The hypergraph is defined as  $\mathcal{G}=(\mathcal{V}, \mathcal{E}, \mathbf{W})$, where $\mathcal{V}=\{v_1, \cdots, v_N\}$  denotes the node set, $\mathcal{E} = \{e_1, \cdots, e_M\}$ denotes the hyperedge set, and $\mathbf{W}$ represents the weight matrix of  the hyperedge set. Here, $N=HW$ and $M$  are the numbers of nodes and hyperedges, respectively. In this paper, we consider each $1 \times 1 \times C$  grid of each feature in $\mathcal{Q}$ as a node. 
We represent the $n$-th node by $\mathbf{f}_n \in \mathbb{R}^{1 \times C}$ and thus all nodes are represented by $\mathbf{F} = [\mathbf{f}_1; \cdots; \mathbf{f}_N] \in {\mathbb{R}^{N \times C}}$.

The traditional hypergraph network allows for unrestricted connections among nodes to capture high-order structure information. Hence, it easily suffers from model collapse (i.e.,
the nodes connected by different hyperedges are the same) during hypergraph learning. To mitigate this problem, we introduce a 
whitening operation to project the nodes 
into a spherical distribution. In fact, 
 the whitening operation plays the role of ``scattering'' on the nodes, thereby
 avoiding 
 the collapse of diverse high-order connections to a single connection. 
 This enables us to explore various high-order relationships across these features effectively.

The whitened node $\mathbf{f}_n^{\prime}$ can be obtained as
\begin{equation} \label{E2}
\mathbf{f}_n^{\prime}= \mathbf{\gamma}_{n} (\mathbf{\sigma}^{-1}  ({\mathbf{f}_n^{\mathrm{T}}}-{\mu_{\mathbf{F}}^{\mathrm{T}}} ))^{{\mathrm{T}}}+\mathbf{\beta}_{n}{,}
\end{equation}
where {$\mathbf{\sigma} \in \mathbb{R}^{C \times C}$} denotes the lower triangular matrix, which is obtained by the Cholesky decomposition
$\mathbf{\sigma \sigma}^{\mathrm{T}} = \frac{1}{N-1}({\mathbf{F}}-{{\mathbf{1}}}\mu_\mathbf{F} )^{{{\mathrm{T}}}}({\mathbf{F}}-{{\mathbf{1}}}\mu_\mathbf{F}) $; {$\mu_\mathbf{F} \in \mathbb{R}^{1\times C}$} denotes the mean vector of $\mathbf{F}$; {${\mathbf{1}} \in \mathbb{R}^{N \times 1}$ is a column vector of all ones}; {$\gamma_n\in \mathbb{R}^{1\times 1}$} and {$\beta_{n}  \in \mathbb{R}^{1\times C}$} are the affine parameters learned from the network. In this way, all the whitened nodes can be represented by  $\mathbf{F^{\prime}} = [\mathbf{f}_1^{\prime}{;}\cdots{;}\mathbf{f}_N^{\prime}] \in {\mathbb{R}^{N\times C}}$.

Similar to \cite{higham2022mean}, we use cross-correlation to 
learn the incidence matrix $\mathbf{H} \in \mathbb{R}^{N\times M}$, i.e., 
    \begin{equation} \label{E3}
\mathbf{H}= \varepsilon ( \Psi(\mathbf{F^{\prime}})\Lambda(\mathbf{F^{\prime}})\Psi(\mathbf{F^{\prime}})^{\mathrm{T}}\Omega(\mathbf{F}^{\prime})),
    \end{equation}
where $\Psi(\cdot)$ represents the linear transformation; $\Lambda (\cdot)$ and  $\Omega (\cdot)$  are responsible for learning a distance metric by a diagonal operation and determining the contribution of the node to the corresponding hyperedges through the learnable parameters, respectively; $\varepsilon (\cdot)$ is the step function.

Based on the learned $\mathbf{H}$, we adopt a hypergraph convolutional operation to aggregate high-order structure information and then enhance feature representations. The relation enhanced feature {${\mathbf{R}} \in  {\mathbb{R}^{ N \times C}}$} can be obtained as
\begin{equation} \label{E4}
{\mathbf{R}}= (\mathbf{I} - \mathbf{D}^{ 1/2}\mathbf{HWB}^{-1}\mathbf{H}^{\mathrm{T}}\mathbf{D}^{-1/2})\mathbf{F}^{\prime} \Theta + \mathbf{F},
    \end{equation}
where {$\mathbf{I} \in {\mathbb{R}^{ N \times N }}$} is the identity matrix;
{$\mathbf{W} \in {\mathbb{R}^{ M \times M}}$ denotes the weight matrix;  
{$\mathbf{D} \in {\mathbb{R}^{ N \times N}}$}  and {$\mathbf{B} \in {\mathbb{R}^{M \times M}}$} represent the node degree matrix and the hyperedge degree matrix {obtained by the broadcast operation}, respectively; {{$\Theta \in {\mathbb{R}^{ C \times C }} $}} denotes the learnable parameters.

Following the above process,  we pass features in ${\mathcal{Q}}$ through the HSL module and obtain a relation-enhanced feature set ${\mathcal{R}}=\{ 
{\mathbf{R}}^{vis}_{L}, {\mathbf{R}}^{vis}_{S},  {\mathbf{R}}^{ir}_{L}, {\mathbf{R}}^{ir}_{S} \}$, where each feature in ${\mathcal{R}}$ is obtained by Eq.~(\ref{E4}).

\subsection{Common Feature Space Learning (CFL) Module}
Conventional feature-level VI-ReID methods usually learn a common feature space based on a contrastive-based loss, which directly minimizes the distances between VIS and IR features.  However, such a manner cannot achieve a reasonable common feature space because of the large modality discrepancy. To address the above problem, it is desirable to learn the middle features from VIS and IR features, enabling us to obtain a  reasonable common feature space. 

A straightforward way to obtain a middle feature is to add or concatenate the VIS or IR features from different ranges. However, the above way cannot generate reliable middle features due to feature misalignment and loss of semantic information.  Therefore, we propose a CFL module, which aligns the features from different modalities and ranges by graph attention (GAT) \cite{guo2021graph} 
and generates reliable middle features, as shown in Figure \ref{F2}.

Specifically, we align each  
feature in $\mathcal{R}$
with the other three features 
in $\mathcal{R}$ and generate a middle feature, which involves the information from different modalities and ranges. We take the alignment between two features ${\mathbf{R}}_{L}^{vis}$ and ${\mathbf{R}}_{S}^{ir}$  as an example.  First, we establish the similarity  between   ${\mathbf{R}}_{L}^{vis}$ and ${\mathbf{R}}_{S}^{ir}$ by using the inner product and the softmax function. This process can be formulated as
	\begin{equation} \label{E5}
		{\mathbf{P}} ={\rm Softmax }(( \theta_{q}{{\mathbf{R}}_{L}^{vis}}) ( \theta_{k}{{\mathbf{R}}_{S}^{ir}})^{\mathrm{T}}),	\end{equation}
where $\theta_{q}$ and $\theta_{k}$ are linear transformations; {$\mathbf{P} \in {\mathbb{R}^{ N \times N}}$} denotes the similarity matrix; {and} $\mathrm{Softmax}(\cdot)$ denotes the softmax function. 

Then, we adopt graph attention to perform alignment between  ${\mathbf{R}}_{L}^{vis}$ and ${\mathbf{R}}_{S}^{ir}$  according to the similarity matrix. Therefore, the aggregated node $\bar{\mathbf{R}}_{ir,S}^{vis,L} {\in {\mathbb{R}^{ N \times C}}}$ is
	\begin{equation} \label{E6}
		\begin{split}
			\bar{\mathbf{R}}_{ir,S}^{vis,L} = &\
			{\rm GAT}({\mathbf{R}}_{L}^{vis},{\mathbf{R}}_{S}^{ir}) \\  = &\ {\rm ReLU }(  \mathbf{P} - \lambda {\rm Mean }(\mathbf{P}){{\mathbf{1}}{\mathbf{1}}^{\mathrm{T}}})   (\theta_{v}{{\mathbf{R}}_{S}^{ir}}),
		\end{split}
	\end{equation}
where  $\rm GAT$($\cdot$) denotes the graph attention {operation; }
$\theta_{v}$ is the linear transformation; $\lambda$ is the balancing parameter that reduces nodes with low {similarity;}  {${\mathbf{1}\mathbf{1}^{\mathrm{T}}} \in {\mathbb{R}^{ N \times N}}$ is a matrix of all {ones; and} ${\rm ReLU}(\cdot)$ and  ${\rm Mean}(\cdot)$ represent the ReLU activation function and the mean operation, respectively.}

Based on the above, a middle feature {${\mathbf{M}}_{L}^{vis} \in {\mathbb{R}^{ N \times C}}$} is obtained by 
aligning ${\mathbf{R}}_{L}^{vis}$ with  ${\mathbf{R}}_{S}^{ir}$,  ${\mathbf{R}}_{L}^{ir}$, and 
${\mathbf{R}}_{S}^{vis}$, that is,
    \begin{equation} \label{E7}
    \begin{split}
{\mathbf{M}}_{L}^{vis}  = &\ {\rm GAT}({\mathbf{R}}_{L}^{vis},{\mathbf{R}}_{S}^{ir})+{\rm GAT}({\mathbf{R}}_{L}^{vis},{\mathbf{R}}_{L}^{ir})+ \\
&\ {\rm GAT}({\mathbf{R}}_{L}^{vis},{\mathbf{R}}_{S}^{vis})+{\mathbf{R}}_{L}^{vis}.
\end{split}
\end{equation}

Similar to Eq. (\ref{E7}), we can get the other reliable middle features. Hence, we obtain the middle feature set ${ \mathcal{\tilde {R}}}=\{ 
{\mathbf{M}}^{vis}_{L}, {\mathbf{M}}^{vis}_{S},  {\mathbf{M}}^{ir}_{L}, {\mathbf{M}}^{ir}_{S} \}$. 
To learn compact feature representations, following previous works \cite{ye2021deep, liu2020parameter}, we apply the holistic and partial generalized mean pooling to each feature in ${ \mathcal{\tilde {R}}}$ and concatenate the pooling features to obtain the 1D middle features. In this way, we can get the 1D middle feature set ${{ \mathcal{\tilde {R}}^{\prime}}}=\{ 
{{\mathbf{m}}^{vis}_{L}, {\mathbf{m}}^{vis}_{S}, \mathbf{m}^{ir}_{L}, 
{\mathbf{m}}^{ir}_{S}\}}$. 
Analogously, we apply the same pooling and concatenation operations to each feature in $\mathcal{R}$ and  obtain 1D feature set $\mathcal{R}'= \{ 
{\mathbf{r}}^{vis}_{L}, {\mathbf{r}}^{vis}_{S},  {\mathbf{r}}^{ir}_{L}, {\mathbf{r}}^{ir}_{S} \}$.

\begin{figure}[t]
\centering
\includegraphics[width=1.0 \columnwidth]{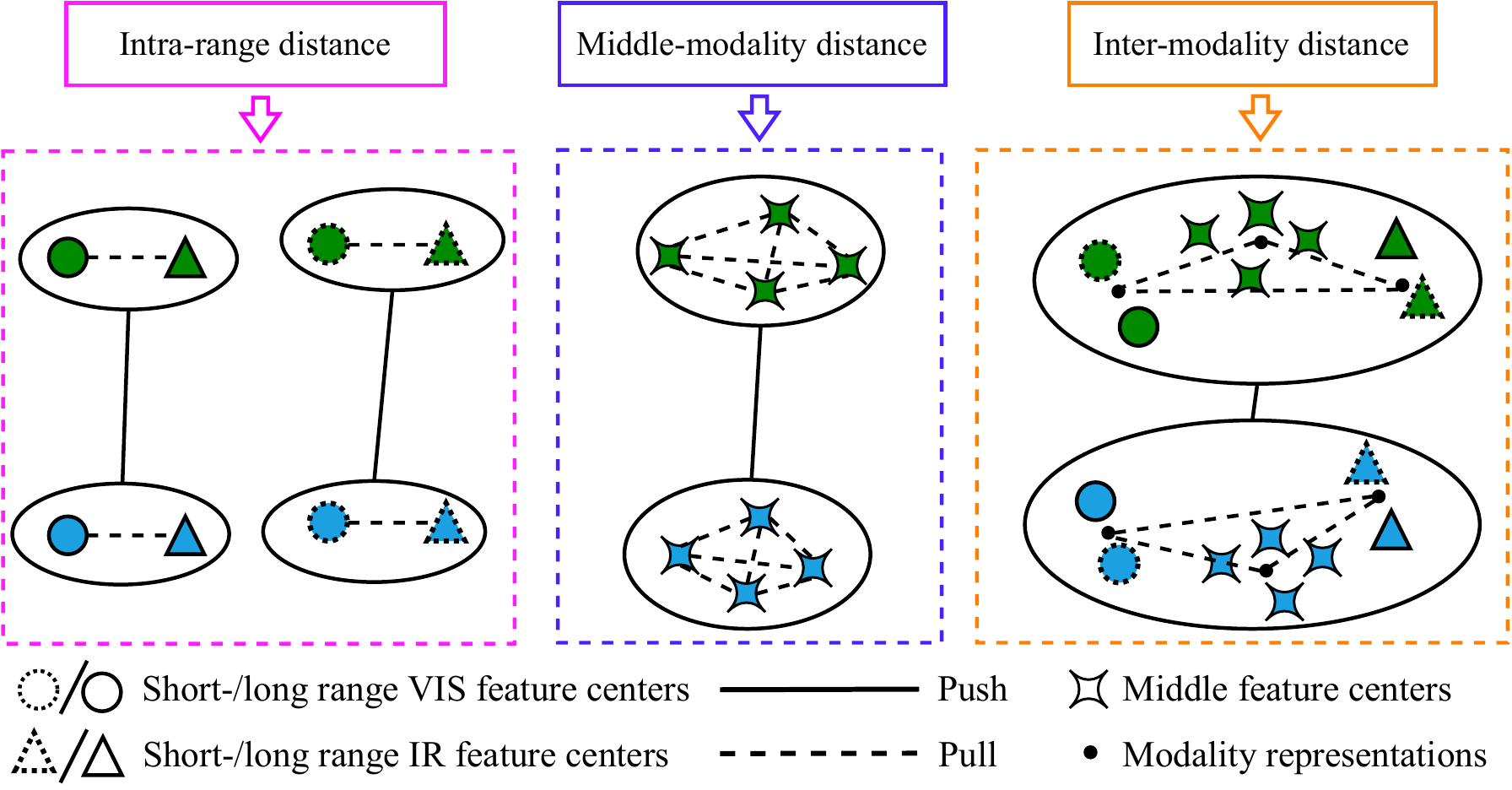}
\caption{Illustration of the proposed MRIC loss. {Different colors represent different identities.}}
\label{F3}
\end{figure}

To mitigate the intra-class {difference} and inter-class discrepancy, we propose the MRIC loss to {improve feature representations and } reduce the distances between the VIS, IR, and middle features. 
The MRIC loss consists of three items: 
an intra-range loss,  a  middle feature loss, and an inter-modality loss based on the identity centers. The illustration of the MRIC loss is shown in Figure \ref{F3}.

 Technically, we first obtain identity centers, which are robust to pedestrian appearance changes, by the weighted average of the features of each person at one modality and a specific range. For instance, the center of the relation-enhanced features for the person with the identity $i$ at the VIS modality and long-range can be obtained as
\begin{equation} \label{E8}
{\mathbf{c}}_{L,i}^{vis}=\sum\limits_{j=1}^K\frac{\exp(\sum_{k=1}^K{{\mathbf{r}}_{L,i,j}^{vis}{\mathbf{r}}_{L,i,k}^{vis}}^{\mathrm{T}})}{\sum_{j=1}^K\exp(\sum_{k=1}^K{{\mathbf{r}}_{L,i,j}^{vis}{\mathbf{r}}_{L,i,k}^{vis}}^{\mathrm{T}})}{\mathbf{r}}_{L,i,j}^{vis},
\end{equation}
where $K$ is the number of VIS features of each person; ${\mathbf{r}}_{L,i,k}^{vis} {\in {\mathbb{R}}^{1 \times C^{\prime}}} $ denotes the $k$-th 1D relation-enhanced long-range VIS feature defined in $\mathcal{R}'$ with the identity $i$. 

Accordingly, we can obtain
the identity center sets $\mathcal{C}_{L}^{vis}$ ($\{{\mathbf{c}}_{L,i}^{vis}\}_{i=1}^{P}$), $\mathcal{C}_{S}^{vis}$ 
 ($\{{\mathbf{c}}_{S,i}^{vis}\}_{i=1}^{P}$), $\mathcal{C}_{L}^{ir}$ ($\{{\mathbf{c}}_{L,i}^{ir}\}_{i=1}^{P}$), $\mathcal{C}_{S}^{ir}$ 
 ($\{{\mathbf{c}}_{S,i}^{ir}\}_{i=1}^{P}$), $\tilde{\mathcal{C}}_{L}^{vis}$ ($\{\tilde{\mathbf{c}}_{L,i}^{vis}\}_{i=1}^{P}$), $\tilde{\mathcal{C}}_{S}^{vis}$ ($\{\tilde{\mathbf{c}}_{S,i}^{vis}\}_{i=1}^{P}$), $\tilde{\mathcal{C}}_{L}^{ir}$ ($\{\tilde{\mathbf{c}}_{L,i}^{ir}\}_{i=1}^{P}$), $\tilde{\mathcal{C}}_{S}^{ir}$ ($\{\tilde{\mathbf{c}}_{S,i}^{ir}\}_{i=1}^{P}$), 
where ${\mathcal{C}}$ and $\tilde{\mathcal{C}}$ represent the center set for the enhanced features and the middle features at a specific range and modality, respectively;
$P$ is the number of person identities in the training set.

The intra-range loss $\mathcal{L}_{MRIC}^{SL}$ is to reduce the distances between the same-range VIS and IR features from the same persons while enlarging the distances between the same-range VIS and IR features from different persons, that is,
    \begin{equation} 
    \begin{split}
    \mathcal{L}_{MRIC}^{SL} =\mathcal{L}_{MRIC}^{{\mathcal{C}_{S}^{vis},\mathcal{C}_{S}^{ir}}}+\mathcal{L}_{MRIC}^{{\mathcal{C}_{L}^{vis},\mathcal{C}_{L}^{ir}}},     
    \label{E9} 
\end{split}
\end{equation}
where 
\begin{equation} 
\small	\begin{split}
		&\mathcal{L}_{MRIC}^{\mathcal{A},\mathcal{B}}=\ -\sum_{i=1}^{P} {\log}\frac{{ \rm exp}{({\mathbf{S}}_{i,i}^{\mathcal{A},\mathcal{B}})}}{\sum_{z=1}^{P} { \rm exp}{({\mathbf{S}}_{i,z}^{\mathcal{A},\mathcal{B}})}} \\   &\ -\sum_{i=1}^{P} {\log}\frac{{ \rm exp}{({\mathbf{S}}_{i,i}^{\mathcal{A},\mathcal{B}})}}{\sum_{z=1}^{P} { \rm exp}{({\mathbf{S}}_{z,i}^{\mathcal{A},\mathcal{B}})}}\ +\sum_{i=1}^{P} {\rm \mathcal{L}_1} {(\mathbf{a}_i - \mathbf{b}_i)}.  \\ 
		\label{E10} 
	\end{split}
\end{equation}
Here, ${\mathbf{S}}^{\mathcal{A},\mathcal{B}} \in \mathbb{R}^{P \times P} $ denotes the cosine similarity matrix between $\mathcal{A}$ and $\mathcal{B}$ (${\mathbf{S}}^{\mathcal{A},\mathcal{B}}_{i,j}$ denotes the cosine similarity between $\mathbf{a}_i$ (the $i$-th element of $\mathcal{A}$) and $\mathbf{b}_j$ (the $j$-th element of $\mathcal{B}$)); ${\mathcal{L}_1 (\cdot)}$ represents  the ${\rm L_1}$ norm. 
By minimizing the first two terms, 
 the similarities of the same person features are enhanced while those of the different person features are reduced. The last term denotes the $L_1$ distance between the same person features. 

The middle-feature loss {$\mathcal{L}_{MRIC}^{MID}$} is to reduce the distances between different middle features, that is,

    \begin{equation} 
    \begin{split}
\mathcal{L}_{MRIC}^{MID} = &\ \mathcal{L}_{MRIC}^{{\tilde{\mathcal{C}}_{{S}}^{vis},\tilde{\mathcal{C}}_{{L}}^{vis}}} + 
                    \mathcal{L}_{MRIC}^{{\tilde{\mathcal{C}}_{S}^{vis},\tilde{\mathcal{C}}_{S}^{ir}}}  +
                    \mathcal{L}_{MRIC}^{{\tilde{\mathcal{C}}_{S}^{vis},\tilde{\mathcal{C}}_{L}^{ir}}}  + \\
                    &\ \mathcal{L}_{MRIC}^{{\tilde{\mathcal{C}}_{L}^{vis},\tilde{\mathcal{C}}_{S}^{ir}}}  + 
                    \mathcal{L}_{MRIC}^{{\tilde{\mathcal{C}}_{L}^{vis},\tilde{\mathcal{C}}_{L}^{ir}}}  + 
                    \mathcal{L}_{MRIC}^{{\tilde{\mathcal{C}}_{S}^{ir},\tilde{\mathcal{C}}_{L}^{ir}}}.        
    \label{E11} 
\end{split}
\end{equation}

The inter-modality loss {$\mathcal{L}_{MRIC}^{VIM}$} is to reduce the intra-class distances and enlarge the inter-class distances {between}  VIS, IR, and middle features, which is expressed as
    \begin{equation} 
    \begin{split}
    \mathcal{L}_{MRIC}^{VIM}= \mathcal{L}_{MRIC}^{\mathcal{C}^{vis},C^{ir}} + \mathcal{L}_{MRIC}^{\mathcal{C}^{vis},C^{mid}} + \mathcal{L}_{MRIC}^{\mathcal{C}^{ir}
    ,\mathcal{C}^{mid}}{,}
    \label{E12} 
\end{split}
\end{equation}
where $\mathcal{C}^{vis}$, $\mathcal{C}^{ir}$, and $\mathcal{C}^{mid}$ denote the identity center sets corresponding to VIS, IR, and middle features, respectively; $\mathcal{C}^{vis}$ and $\mathcal{C}^{ir}$ are obtained by averaging all the features from the same modality for each person; $\mathcal{C}^{mid}$ is obtained by averaging all the middle features for each person.

Therefore, the MRIC loss is
    \begin{equation}
	\begin{split}
		\mathcal{L}_{MRIC}= \mathcal{L}_{MRIC}^{SL}+\mathcal{L}_{MRIC}^{MID}+\mathcal{L}_{MRIC}^{VIM}.
\label{E13} 
\end{split}
\end{equation}

\subsection{Joint Loss}

 The joint loss is defined as
    \begin{equation}
    \begin{split}
    \mathcal{L}= \mathcal{L}_{CE}+\mathcal{L}_{TRI}+\mathcal{L}_{MRIC},
    \label{E15} 
    \end{split}
    \end{equation}
where $\mathcal{L}_{CE}$ represents the 
  cross-entropy loss  and $\mathcal{L}_{TRI}$ denotes the triplet loss \cite{hermans2017defense}.

The training of HOS-Net is given in \textit{Supplement A}.

\section{Experiments}
\begin{table*}[!t]
    \begin{center}
    \small
    {
    \begin{tabular}{l| c |c |c |c |c |c }
    \hline
       
    \multirow{3}*{Methods} & \multicolumn{2}{c|}{SYSU-MM01}& \multicolumn{2}{c|}{RegDB}& \multicolumn{2}{c}{LLCM}\\
    \cline{2-7}
    ~ & \multicolumn{1}{c|}{All search}& \multicolumn{1}{c|}{Indoor search}& \multicolumn{1}{c|}{VIS to IR}& \multicolumn{1}{c|}{IR to VIS}& \multicolumn{1}{c|}{VIS to IR}& \multicolumn{1}{c}{IR to VIS}\\
    \cline{2-7}
    ~ & R-1 / mAP&R-1 / mAP&R-1 / mAP&R-1 / mAP&R-1 / mAP&R-1 / mAP\\
    \hline
        D$^{2}$RL \cite{wang2019learning}  &  28.9 / 29.2    &-    /   - &   43.4 /    44.1  &   - /   -&-    /   - &-    /   -\\
        Hi-CMD \cite{choi2020hierarchical}  &  34.9 / 35.9    &-    /   - &   70.9 /    66.0  &   - /   -&-    /   - &-    /   -\\
        JSIA-ReID \cite{wang2020cross}  &  38.1  / 36.9    &43.8    /    52.9 &   48.1 /   48.9  &   48.5 /  49.3&-    /   - &-    /   -\\
        X-Modality \cite{li2020infrared} &  49.9 / 50.7    &-  / - &   62.2    /   60.2  &   - /   -&-    /   -&-    /   -\\
        DDAG \cite{ye2020dynamic}  &  54.8 / 53.0    &61.0 / 68.0 &69.3 / 63.5 &68.1 / 61.8 &48.0   /  52.3 &  40.3 / 48.4  \\
        LbA \cite{park2021learning}  &  55.4 /  54.1&  58.5 / 66.3 & 74.2 / 67.6&  67.5 / 72.4 &50.8    /   55.6 & 43.8 /   53.1      \\
        G$^{2}$DA \cite{wan2023g2da} &63.9 / 60.7 &71.0 / 76.0 &74.0 / 65.5 &69.7 / 62.0 &-    /   - &-    /   -\\

       TSME \cite{liu2022revisiting}  &  64.2 / 61.2 &64.8 / 71.5 &87.4 / 76.9 &86.4 / 75.7 &-    /   - &-    /   -\\

       SPOT \cite{chen2022structure}& 65.3 / 62.3 &69.4 / 74.6 &80.4 / 72.5 &79.4 / 72.3 &-    /   - &-    /   -\\
 
        PMT \cite{lu2022learning} & 67.5 / 65.0  &71.7 / 76.5 &84.8 / 76.6 &84.2 / 75.1 &-    /   - &-    /   -\\
        CAJ \cite{Ye_2021_ICCV}  &  69.9  / 66.9  &76.3  / 80.4  &85.0 / 79.1  &84.8  / 77.8 &56.5      /    59.8 &  48.8 /   56.6      \\
        MMN \cite{zhang2021towards}  &  70.6 / 66.9 &76.2 / 79.6 & \underline{91.6} / 84.1 &87.5 / 80.5 &59.9     /  62.7 &  52.5  / 58.9     \\

        MAUM \cite{liu2022learning}  &  71.7 / 68.8 &77.0 / 81.9 &87.9 / \underline{85.1} &87.0 / \underline{84.3}&-    /   - &-    /   -\\
    
        DEEN \cite{zhang2023diverse}  &  \underline{74.7} / \underline{71.8} &\underline{80.3} / \underline{83.3} &91.1 / \underline{85.1} &\underline{89.5} / 83.4 &\underline{62.5}    /   \underline{65.8}& \underline{54.9}  / \underline{62.9}    \\
    \hline
        HOS-Net (Ours)  &  \textbf{75.6} / \textbf{74.2}    &\textbf{84.2}    / \textbf{86.7} &   \textbf{94.7} /    \textbf{90.4}  &   \textbf{93.3} /   \textbf{89.2} &\textbf{64.9}       /  \textbf{67.9}&  \textbf{56.4} / \textbf{63.2}     \\
    \hline
    \end{tabular}
}
    \caption{Comparisons with state-of-the-art methods on the SYSU-MM01, RegDB and LLCM datasets. The bold font 
    and the underline denote the best and second-best performance, respectively.}
        \label{T1}

\end{center}
\end{table*}

\subsection{Experimental Settings}

\noindent \textbf{Datasets.}
The SYSU-MM01 dataset \cite{wu2020rgb} contains a total of 30,071 VIS images and 15,792 IR images from 491 different
 identities. The RegDB dataset \cite{nguyen2017person} consists of 412 identities, where each identity has 10 VIS images and 10 IR images captured by two overlapping cameras. The LLCM dataset \cite{zhang2023diverse} is captured in low-light environments. The training set contains 713 identities (with 16,946 VIS images and 13,975 IR images) while the test set contains 351 identities (with 8,680 VIS images and 7,166 IR images).

\noindent \textbf{Implementation Details.} All the images are resized to 256$\times$128 with horizontal flip, random erasing, and channel augmentation for data augmentation \cite{Ye_2021_ICCV} during the training phase. For each mini-batch, we randomly choose 8 identities, where 4 VIS images and 4 IR images of each identity are selected. We adopt AGW \cite{ye2021deep} as our backbone. We use the warm-up strategy to update the learning rate from 0.01 to 0.1 at the first 10 epochs. At the 20 and 50 epochs, the learning rates are set to 0.01 and 0.001, respectively.  We use SGD as the optimizer and the momentum parameter is set to 0.9.  The total number of training epochs is set to 120. Our proposed HOS-Net is implemented with the PyTorch on an NVIDIA RTX3090 GPU. The number of hyperedges {$M$} is set to 256.  $\lambda$ in Eq. (\ref{E6}) is set to 1.3.

Cumulative Matching characteristics (CMC) and mean Average Precision (mAP) are used as our evaluation metrics. CMC measures the matching
probability of the ground-truth person occurring in the top-$k$ retrieved results (Rank-$k$ accuracy). Besides, we randomly divide the RegDB dataset for training and testing. 
The above process is repeated ten times and we report the average performance. We also randomly split the gallery set of the SYSU-MM01 and LLCM datasets ten times to report the average performance.

\subsection{Comparison with State-of-the-Art Methods}
The comparison results are given in Table  \ref{T1}. More results are shown in \textit{Supplement B}.

\noindent \textbf{SYSU-MM01.} As shown in Table  \ref{T1}, our proposed HOS-Net obtains the best or comparable performance among all the competing methods.  Specifically,  HOS-Net gives about 13.0\%  and 15.2\% improvements in terms of mAP over some image-level methods (such as  JSIA-ReID and TSME) for both all and indoor search modes, respectively. Compared with the CNN-based method (DDAG) and Transformer-based method (PMT), HOS-Net surpasses them by at least 8.1\% in Rank-1 and 9.2\% in mAP for the all search mode. Moreover, for the indoor search mode, HOS-Net outperforms the second-best method DEEN  by 3.9\% in Rank-1 and 3.4\% in mAP. 
DEEN ignores the importance of high-order structure information, leading to inferior performance.

\noindent \textbf{RegDB.} From Table  \ref{T1}, we can also observe that our proposed HOS-Net achieves the best performance for two search modes.  For two search modes, our HOS-Net outperforms MMN 
by 3.1\%/6.3\%  and 5.8\%/8.7\% in Rank-1/mAP, respectively. Moreover, 
compared with  G$^{2}$DA and SPOT, which rely on high-quality person structure labels to obtain modality-shared features, HOS-Net improves the Rank-1 and mAP by at least 13.9\% and 16.9\%, respectively, for the {IR to VIS} search mode. This further indicates the superiority of our high-order structure-based network for VI-ReID.

\noindent \textbf{LLCM.} We also report the comparison results on the LLCM dataset in Table \ref{T1}. For the {IR to VIS} search mode, our HOS-Net outperforms MMN by 3.9\% and 4.3\% in terms of Rank-1 and mAP, respectively. Moreover, HOS-Net performs significantly better than the second-best DEEN for the {VIS to IR} search mode, achieving the best results with 64.9\%/67.9\% in Rank-1/mAP. Therefore, HOS-Net can learn a discriminative and reasonable common feature space to reduce the modality discrepancy.

\subsection{Ablation Studies}

  \begin{table}[!t]
  \begin{center}
    \small
    \vspace{-0.4cm}
    {
    \begin{tabular}{c | l  |c  | c }
    \hline   
    \multirow{2}*{\#}&\multirow{2}*{Methods}&\multicolumn{1}{c|}{SYSU-MM01}&\multicolumn{1}{c}{RegDB}\\
    \cline{3-4}
    ~& ~ &   R-1 / mAP & R-1 / mAP\\
    \hline  
      1 & Baseline & 69.9 / 66.9 & 85.0 / 79.1 \\
      \hline
      2 & Baseline+SLE & 71.7 / 69.4 & 89.6 / 84.8 \\
      3 & +HSL &  73.3 / 72.4 & 92.0 / 87.1\\
      4 & +CFL &72.1 / 70.2 & 91.6 / 86.5 \\
      5 & +HSL+CFL & 74.0  / 72.9 &  92.7 / 87.8  \\
      {6} & +CFL +$\mathcal{L}_{MRIC}$&74.5  / 72.7 &   93.2 / 88.4  \\
      7 &  +HSL+CFL +$\mathcal{L}_{MRIC}$ & \textbf{75.6} / \textbf{74.2} & \textbf{94.7} / \textbf{90.4}\\

    \hline 
    \end{tabular}
    }
    \caption{The influence of key components of HOS-Net on the performance on the SYSU-MM01 and RegDB datasets.}
        \label{T3}

    \end{center}
    
    \end{table}

\noindent \textbf{Effectiveness of Key Components.} We conduct ablation studies to validate the effectiveness of each key component of the proposed HOS-Net (including SLE, HSL, CFL, and the MRIC loss). The results are shown in Table  \ref{T3}, where Method 1 represents the baseline AGW method.

\noindent \textbf{SLE:} 
By introducing SLE, Method 2 achieves about 2.5\% and 5.7\% higher mAP than Method 1 on the SYSU-MM01 and RegDB datasets, respectively. This shows the effectiveness of our SLE, which explores different ranges of person features by taking advantage of both CNN and Transformer. \textbf{HSL:} By incorporating HSL into Method 2, Method 3 achieves 
1.6\%/3.0\% and 2.4\%/2.3\% improvements in Rank-1/mAP on two datasets, respectively. This validates the importance of HSL, which adopts the whitened hypergraph network to model the high-order relationship across different local features of each person image and avoid model collapse. \noindent \textbf{CFL:} Method 5 introduces CFL to Method 3 and it  obtains higher accuracy (0.7\%/0.7\% improvements in Rank-1/mAP on the RegDB dataset) than Method 3.   This demonstrates that learning reliable middle features can effectively reduce the modality discrepancy. \textbf{The MRIC loss:}  Compared with Method 5, Method 7 achieves 1.6\%/1.3\% and 2.0\%/2.6\% improvements in Rank-1/mAP on two datasets, respectively. The MRIC loss can improve feature representations and reduce discrepancies between the VIS and IR modalities, achieving a reasonable common feature space.

  \begin{table}[t]
  
  \begin{center}
    \small
    {
    \begin{tabular}{ c c |c |  c}
    \hline   
  \multicolumn{2}{c|}{Settings}&\multicolumn{1}{c|}{SYSU-MM01}&\multicolumn{1}{c}{RegDB}\\
  \hline 
    Hypergraph & {Whitening}  &  R-1 / mAP & R-1 / mAP\\
    \hline  
     - & -  & 71.7 / 69.4 & 89.6 / 84.8 \\
      \checkmark & - & 72.5  / 70.3 & 91.1 / 86.3  \\
      \checkmark & \checkmark   & \textbf{73.3} / \textbf{72.4} & \textbf{92.0} / \textbf{87.1}\\

    \hline 
    \end{tabular}
    }
    \caption{The influence of the hypergraph and the whitening operation on the SYSU-MM01 and RegDB datasets.}
    \label{T5}
    \end{center}  
    \end{table}

\noindent \textbf{Effectiveness of the Hypergraph and the Whitening Operation.}
HSL is based on a whitened hypergraph network to discover the high-order relationship of person features and avoid model collapse. As shown in Table \ref{T5}, by modeling the high-order structure with the hypergraph, the model brings about 0.8\%/0.9\% gains in Rank-1/mAP on SYSU-MM01. Note that the original hypergraph network allows unrestricted connections among nodes to capture high-order structure information, suffering from model collapse during hypergraph learning. 
By adding the whitening operation into the hypergraph learning, the performance is improved by 0.8\%/2.1\% and 0.9\%/0.8\% in Rank-1/mAP on two datasets, respectively. Hence, the whitening operation is beneficial to alleviate mode collapse and improve the final performance.

 \begin{table}[t]
  \begin{center}
    \small
    \vspace{-0.5cm}
    {
    \begin{tabular}{ c c c |c  }
    \hline   
  \multicolumn{3}{c|}{Settings}&\multicolumn{1}{c}{SYSU-MM01}\\
  \hline   
    ~ & Modalities & Ranges   &   R-1 / mAP \\
    \hline
  \multirow{3}*{{Addition}} & \checkmark & -  & 72.0  / 71.1  \\
     ~ & -  & \checkmark  &71.8 / 70.9  \\
      ~ & \checkmark  & \checkmark  & 71.4 / 70.4
       \\
    \hline   

  \multirow{3}*{Concatenation} & \checkmark & - & 71.9  / 70.8 \\
     ~ & -  & \checkmark  &72.3 / 71.1  \\
      ~ & \checkmark  & \checkmark  & 72.5 / 71.2 \\
    \hline   

  \multirow{3}*{GAT} & \checkmark & -  & 73.4 / 72.3 \\
     ~ & -  & \checkmark  &  73.6 / 72.4 \\
      ~ & \checkmark  & \checkmark  & \textbf{74.0} / \textbf{72.9}  \\
    \hline   
    \end{tabular}
    }
    \caption{The influence of generating middle features from different modality and range features on SYSU-MM01.}
    \label{T6}
    \end{center}
    
    \end{table}

\noindent \textbf{Influence of Different Middle Features.} The CFL module leverages graph attention to align features from different modalities and ranges to generate reliable middle features. In this subsection, we evaluate the influence of different middle features on the performance. The results are given in Table  \ref{T6}. Compared with the methods that generate 
middle features by adding or concatenating the VIS or IR features, our method with GAT improves mAP on the SYSU-MM01 datasets, respectively. This clearly indicates the effectiveness of the middle features generated by GAT.

\noindent \textbf{Influence of Each Term in the MRIC Loss.} The MRIC loss is proposed to improve feature representations and reduce the distances between the VIS, IR, and middle features. As shown in Table  \ref{T7}, when all the terms in the MRIC loss are used to jointly train the network, Rank-1/mAP is improved by 1.6\%/1.3\% and 2.0\%/2.6\% in comparison with HOS-Net trained without the MRIC loss on two datasets, respectively. This indicates that HOS-Net trained with the MRIC loss can achieve a reasonable common feature space.

 \begin{table}[t]
  \begin{center}
    \small
    \vspace{-0.2cm}
    {
    \begin{tabular}{ l  |c | c }
    \hline   
  \multirow{2}*{Settings}&\multicolumn{1}{c|}{SYSU-MM01}&\multicolumn{1}{c}{RegDB}\\
  \cline{2-3}   
    ~  &  R-1 / mAP & R-1 / mAP\\
  \hline
   -  & 74.0 / 72.9 & 92.7 / 87.8 \\
   +$\mathcal{L}_{MRIC}^{SL}$  & 74.3  / 73.3 & 93.4 / 88.6  \\
   +$\mathcal{L}_{MRIC}^{MID}$  & 74.4  / 73.2 & 93.3 / 88.3  \\
    +$\mathcal{L}_{MRIC}^{SL}$ +$\mathcal{L}_{MRIC}^{MID}$ & 75.0  / 73.8 & 93.8 / 89.2 \\
   +$\mathcal{L}_{MRIC}^{SL}$ +$\mathcal{L}_{MRIC}^{MID}$+$\mathcal{L}_{MRIC}^{VIM}$ & \textbf{75.6} / \textbf{74.2} & \textbf{94.7}  / \textbf{90.4} \\

    \hline   
    \end{tabular}
    }    
    \caption{The influence of each term in the MRIC loss.}
    \label{T7}

    \end{center}
    \end{table}

\begin{figure}[t]
\centering
\includegraphics[width=1.0\columnwidth]{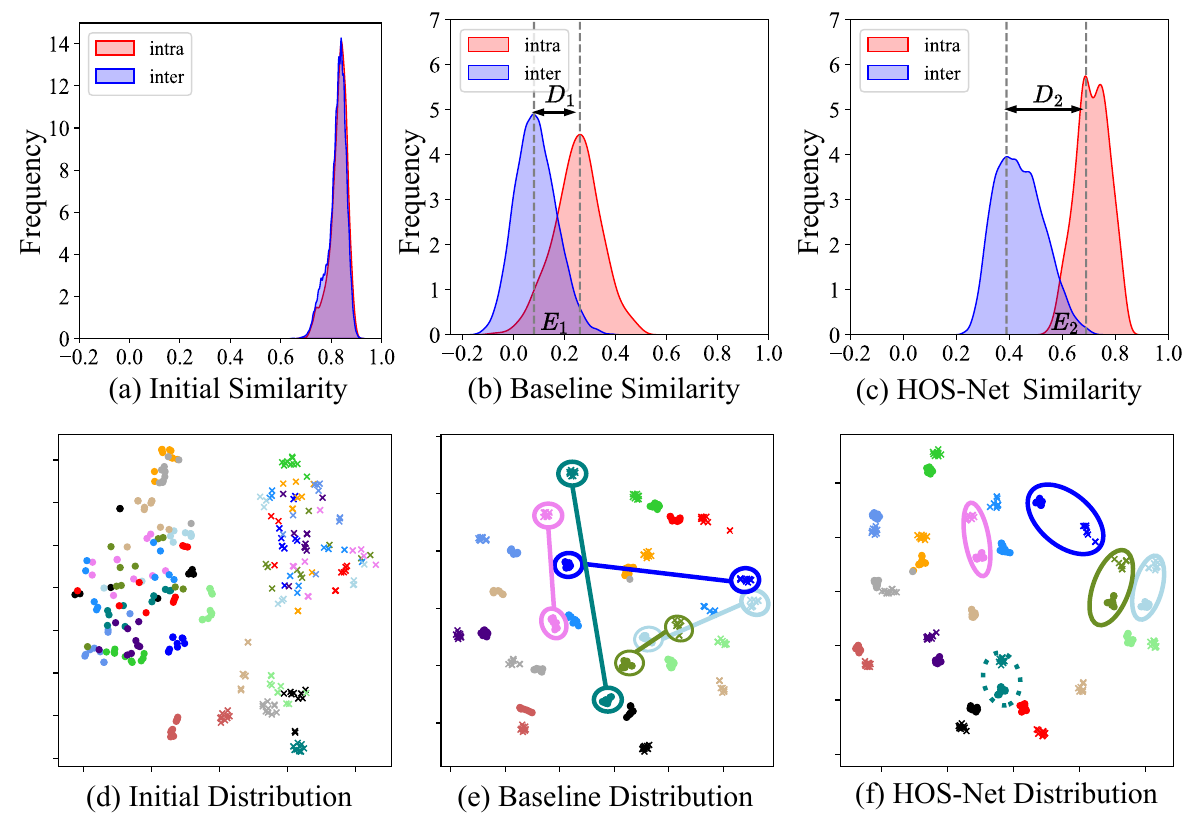}
\caption{ (a-c) give the distributions of the intra-class and inter-class similarities of VIS and IR modality features on the SYSU-MM01 dataset.
(d-f) visualize the distribution of person features from VIS and IR modalities in the 2D feature space on the SYSU-MM01 dataset. Circles and crosses represent features from VIS and IR modalities, respectively.}
\label{F4}
\end{figure}

\noindent \textbf{Feature Distribution Visualization.} 
 We randomly sample { a total of 40,000 positive and negative matching pairs } from the test set and visualize the cosine similarity distribution on the SYSU-MM01 dataset, as shown in Figure \ref{F4}(a-c).
 The cosine similarity of positive matching pairs is increased while the similarity differences between positive and negative matching pairs are enlarged.  

This shows that the proposed HOS-Net can effectively reduce the modality gap between VIS and IR modalities. Then, we adopt t-SNE \cite{van2008visualizing} to illustrate feature distributions obtained by different methods on the SYSU-MM01 dataset, as shown in Figure \ref{F4}(d-f). Compared with the Baseline, our HOS-Net can reduce intra-class differences and enlarge inter-class discrepancies in different modalities.
More ablation studies and visualization results can refer to \textit{Supplement C and D}.

\section{CONCLUSION}
In this paper, we propose a novel HOS-Net consisting of the backbone, SLE, HSL, and CFL modules for VI-ReID. {The} SLE module is first designed to learn short-range and long-range features by taking advantage of both CNN and Transformer. Then, {the} HSL module exploits diverse high-order structure information of features without suffering from model collapse based on 
a whitened hypergraph. Finally, {the} CFL module generates reliable middle features and obtains a reasonable common feature space. 
Extensive experiments on three VI-ReID benchmarks verify the effectiveness of HOS-Net in comparison with several state-of-the-art methods. 
Currently, the training complexity of our method is still high (see \textit{Supplement C} for more details). We plan to explore new ways to reduce the training complexity.

\section{Acknowledgments}
This work was supported by the National Natural Science Foundation of China under Grants 62372388 and 62071404, by the Open Research Projects of Zhejiang Lab under Grant 2021KG0AB02, and by the Natural Science Foundation of Fujian Province under Grant 2021J011185.
\bibliography{aaai24}

\appendix
\renewcommand{\thefigure}{\Roman{figure}}
\renewcommand{\thetable}{\Alph{table}}
\setcounter{figure}{0} 
\setcounter{table}{0} 
\section{Supplement A. More Details of HOS-Net}

{The whole training process of the proposed HOS-Net is summarized
in Algorithm \ref{alg:algorithm}. Moreover, we also illustrate the proposed $\mathcal{L}_{MRIC}^{\mathcal{A}, \mathcal{B}}$ in Eq. (10) and the pseudo-code of the $\mathcal{L}_{MRIC}^{\mathcal{A}, \mathcal{B}}$ in Figure \ref{MRIC_a_b} and Listing \ref{lst:listing}, respectively.}

\begin{algorithm}[!t]
\caption{The training process of HOS-Net.}
\label{alg:algorithm}
\textbf{Input}: The training set $\mathcal{D}$, the initial parameters ($\theta_{hos}$) of HOS-Net, the number of training epochs $E$, the number of mini-batches $B$, the number of identities $P$ in each mini-batch.\\ 
\textbf{Output}: The final parameters ($\theta_{hos}^{\prime}$) of HOS-Net.\\
\begin{algorithmic}[1]
\FOR{$e=1$ to $E$}
\FOR{each mini-batch $\{\mathbf{I}_{b}^{vis},\mathbf{I}_{b}^{ir} \}_{b=1}^{B}$ in $\mathcal{D}$} 
\FOR{each image in ${\{\mathbf{I}_{b}^{vis}, \mathbf{I}_{b}^{ir}} \}_{b=1}^B$}
\STATE Feed $\mathbf{I}_{b}^{vis}$ and $ \mathbf{I}_{b}^{ir}$ to the backbone and obtain the basic features $\mathbf{B}^{vis}$ and $\mathbf{B}^{ir}$.
\STATE Obtain the feature set $\mathcal{Q}=\{ 
\mathbf{F}^{vis}_{L}, \mathbf{F}^{vis}_{S},  \mathbf{F}^{ir}_{L}, $ $\mathbf{F}^{ir}_{S} \}$ via Eq. (1).
\STATE Apply the whitening operation to project the nodes 
of each feature in $\mathcal{Q}$ into a spherical distribution via Eq. (2).           
\STATE Obtain the relation-enhanced feature set ${\mathcal{R}}=\{ 
{\mathbf{R}}^{vis}_{L}, {\mathbf{R}}^{vis}_{S},  {\mathbf{R}}^{ir}_{L}, {\mathbf{R}}^{ir}_{S} \}$ via Eqs. (3) and (4).
\STATE Align the different modalities and ranges features to get the reliable middle feature set ${ \mathcal{\tilde {R}}}=\{ 
{{\mathbf{M}}^{vis}_{L}, {\mathbf{M}}^{vis}_{S},
\mathbf{M}}^{ir}_{L}, 
{\mathbf{M}}^{ir}_{S}\}$ via Eqs. (5), (6), and (7).
\STATE Apply the holistic and partial generalized mean pooling and concatenation to get the 1D feature representation set (${\mathcal{R}^{\prime}}$ and ${\mathcal{\tilde{R}^{\prime}}}$) of ${\mathcal{R}}$ and ${\mathcal{\tilde{R}}}$.
\ENDFOR
\STATE Obtain the identity center sets $\mathcal{C}_{L}^{vis}$ ($\{{\mathbf{C}}_{L,i}^{vis}\}_{i=1}^{P}$), $\mathcal{C}_{S}^{vis}$ 
 ($\{{\mathbf{C}}_{S,i}^{vis}\}_{i=1}^{P}$), $\mathcal{C}_{L}^{ir}$ ($\{{\mathbf{C}}_{L,i}^{ir}\}_{i=1}^{P}$), $\mathcal{C}_{S}^{ir}$ 
 ($\{{\mathbf{C}}_{S,i}^{ir}\}_{i=1}^{P}$), $\tilde{\mathcal{C}}_{L}^{vis}$ ($\{\tilde{\mathbf{C}}_{L,i}^{vis}\}_{i=1}^{P}$), $\tilde{\mathcal{C}}_{S}^{vis}$ ($\{\tilde{\mathbf{C}}_{S,i}^{vis}\}_{i=1}^{P}$), $\tilde{\mathcal{C}}_{L}^{ir}$ ($\{\tilde{\mathbf{C}}_{L,i}^{ir}\}_{i=1}^{P}$), $\tilde{\mathcal{C}}_{S}^{ir}$ ($\{\tilde{\mathbf{C}}_{S,i}^{ir}\}_{i=1}^{P}$).
\STATE  Compute the joint loss via Eq.~(14).
\STATE Update $\theta_{hos}$ by stochastic gradient descent. 
\ENDFOR
\ENDFOR

\end{algorithmic}
\end{algorithm}

\begin{figure}[!t]
\centering
\includegraphics[width=1.0 \columnwidth]{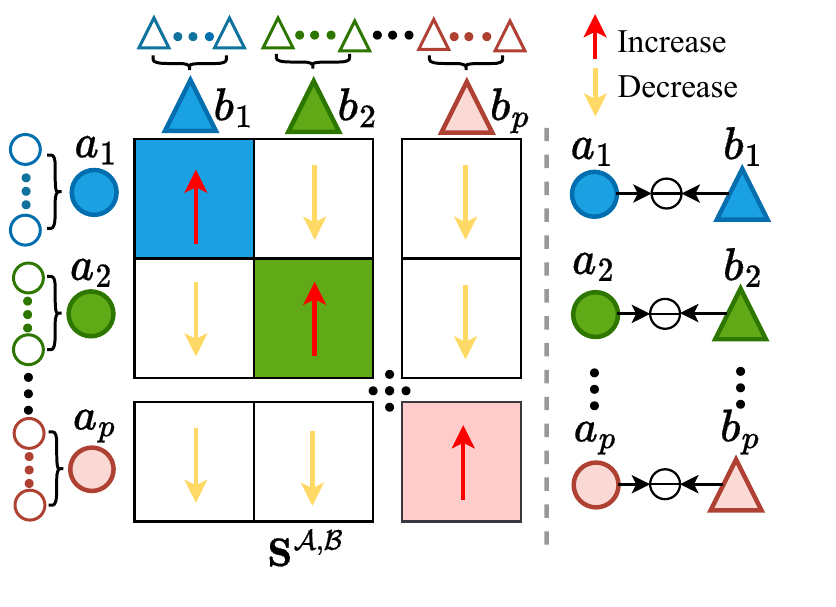}
\caption{ { Illustration of the proposed $\mathcal{L}_{MRIC}^{\mathcal{A},\mathcal{B}}$.}}
\label{MRIC_a_b}
\end{figure}

\begin{listing}[tb]%
\caption{Pseudo code of the MRIC loss.}%
\label{lst:listing}%
\begin{lstlisting}[language=python]
def identity_weight(x): 
    sim = sum(multiplication(x, x.T)
    return softmax(sim).view(4, -1)
    
def get_center(x, i, K):
    center = identity_weight(x[i * K:(i + 1) * K]) * x[i * K:(i +1) * K]
    return sum(center).view(1, -1)
    
# input x1: tensor(B, C')
# input x2: tensor(B, C')
# B: batch-size, B=P*K 
# P: the number of identities, K: the number of features of each identity 
# C': feature dimension
def MRIC(x1, x2, B, K):
    a = []
    b = []
    # B//K = P (identity group, get identity center) 
    for i in range(B//K): 
        # tensor(1, C')
        a.append(get_center(x1,i,K))
        b.append(get_center(x2,i,K)) 
    # cosine simility matrix (S) of center a and center b
    s = multiplication(l2_norm(a), l2_norm(b).T) 
    # pseudo label [0, ..., B//K-1] 
    labels = arange(B//K) 
    # the fisrt term of L_{MRIC}^{a, b}
    a_2_b  = cross_entropy(s, labels) 
    # the second term of L_{MRIC}^{a, b}
    b_2_a = cross_entropy(s.T, labels) 
    # the third term of L_{MRIC}^{a, b}
    l1_distance = mean(l1_norm((a - b)) 
    return  a_2_b+ b_2_a+l1_distance
\end{lstlisting}
\end{listing}

\section{Supplement  B. More Comparison Results}

We show comparisons with several state-of-the-art methods under the multi-shot setting of the SYSU-MM01 dataset, as shown in Table \ref{T-MS}. Our HOS-Net outperforms the second-best MPANet by 6.1\% and 7.9\%  in terms of mAP in all search and indoor search, respectively. These results further demonstrate that our HOS-Net can effectively mitigate the modality discrepancy between VIS and IR modalities and improve the VI-ReID performance.

\section{Supplement C. More Ablation Study Results}

\begin{figure}[!t]
\centering
\includegraphics[width=1.0\columnwidth]{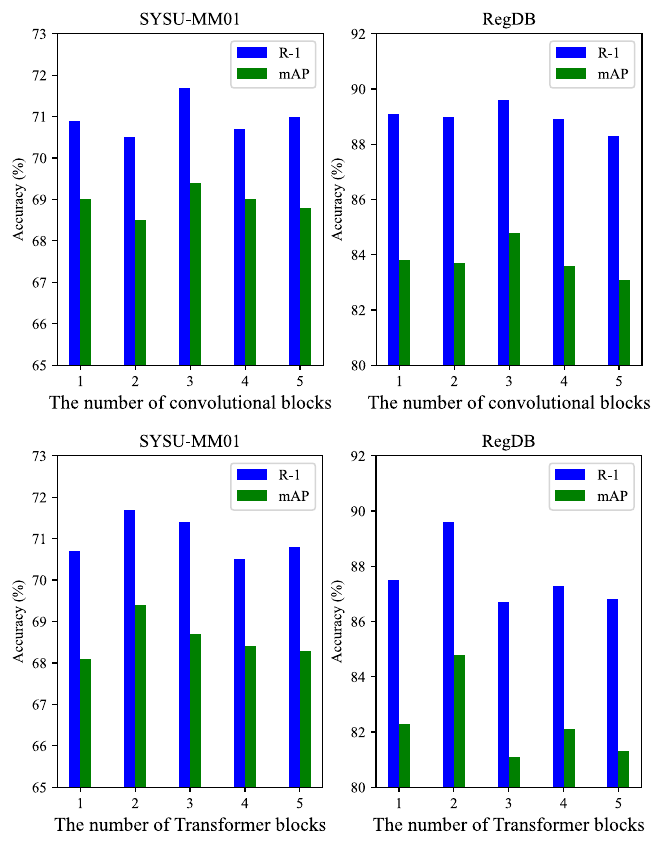}
\caption{{The influence of the number of convolutional and Transformer blocks on the SYSU-MM01 and RegDB datasets.}}
\label{Number}
\end{figure}

\begin{table}[!t]
    \begin{center}
    \small
    {
    \begin{tabular}{l| c |c } 
    \hline
      
    \multirow{3}*{Methods} & \multicolumn{2}{c}{SYSU-MM01} \\
    \cline{2-3}
    ~ & \multicolumn{1}{c|}{All search }& \multicolumn{1}{c}{Indoor search}\\
    \cline{2-3}
    ~ & R-1 / mAP&R-1 / mAP \\
  \hline
  Zero-Pad \cite{wu2017rgb-1}& 19.1  / 10.9 & 24.4 / 18.9\\
  AlignGAN \cite{wang2019rgb-1} &  51.5 / 33.9 & 57.1  / 45.3\\
  cm-SSFT \cite{lu2020cross-1}& 63.4 / 62.0 & 73.0 / 72.4\\
  CM-NAS \cite{fu2021cm-1} & 68.7 / 53.5 & 76.5 / 65.1\\
  TSME \cite{liu2022revisiting-1} & 70.3 / 54.4 & 76.8 /  65.0\\
  FMCNet \cite{zhang2022fmcnet-1} & 73.4 / 56.1 & 78.9 / 63.8\\
  MPANet \cite{wu2021discover-1}&\underline{75.6} / \underline{62.9} & \underline{84.2} / \underline{75.1}\\
    HOS-Net (Ours)  &  \textbf{79.7}  / \textbf{69.0}    &\textbf{89.8}    /   \textbf{83.0} \\
  \hline
    \end{tabular}
}
\caption{{Comparisons with state-of-the-art methods on the SYSU-MM01 dataset (multi-shot setting). The bold font and the underline denote the best and the second-best performance, respectively.}}
\label{T-MS}

\end{center}
\end{table}

 \begin{figure}[!t]
\centering
\includegraphics[width=1.0\columnwidth]{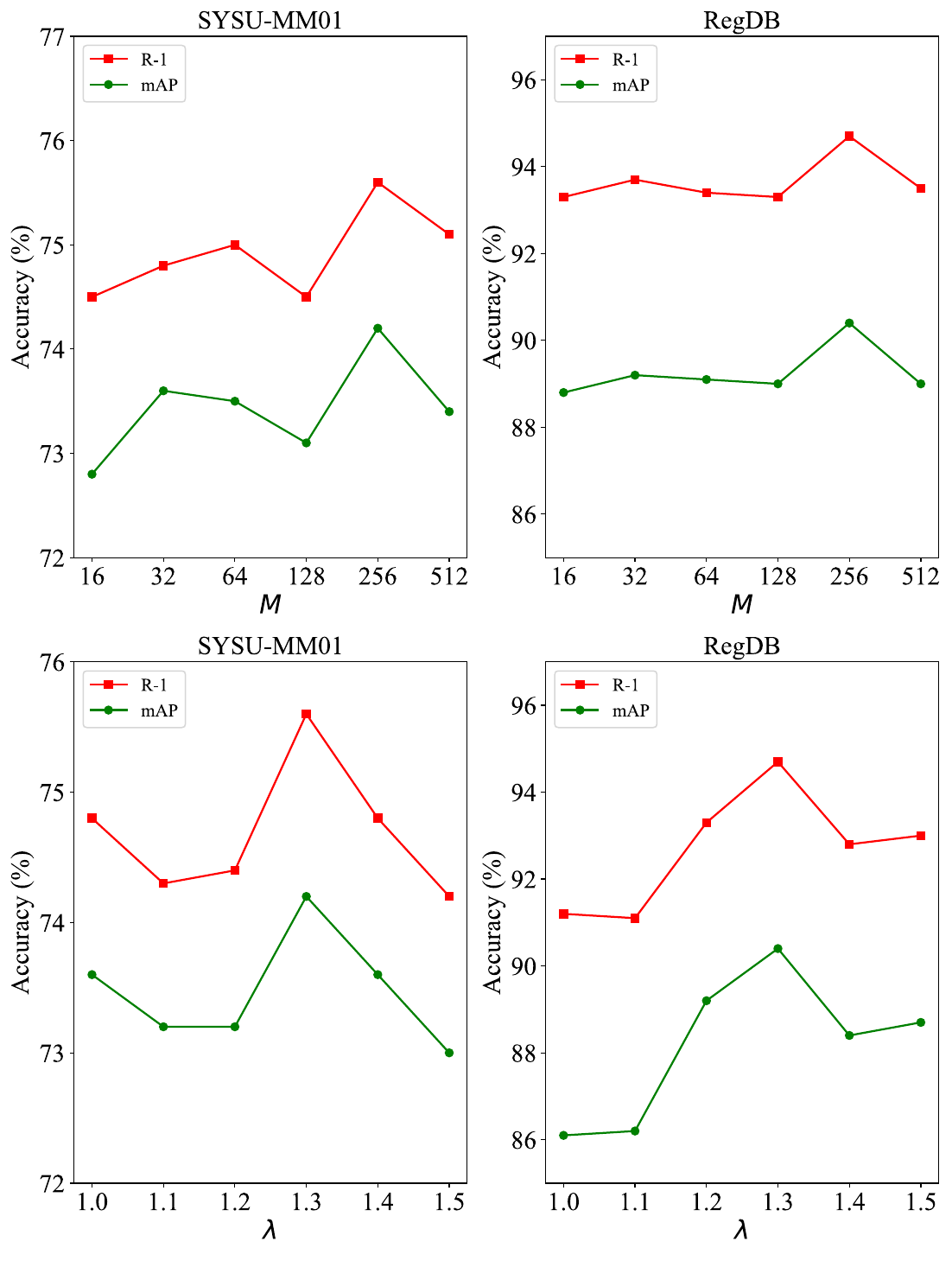}
\caption{The influence of the number of hyperedges $M$ and the balancing parameter $\lambda$ in Eq. (6) on the SYSU-MM01 and RegDB datasets.}
\label{Lambda}
\end{figure}

 \begin{table}[!t]
  \begin{center}
    \small
    {
    \begin{tabular}{ c c c |c  }
    \hline  
  \multicolumn{3}{c|}{Settings}&\multicolumn{1}{c}{RegDB}\\
  \hline  
    ~ & Modalities & Ranges   &   R-1 / mAP \\
    \hline
  \multirow{3}*{{Addition}} & \checkmark & -  & 91.1  / 86.4  \\
     ~ & -  & \checkmark  &91.3 / 86.6  \\
      ~ & \checkmark  & \checkmark  & 90.5 / 85.8
       \\
    \hline  

  \multirow{3}*{Concatenation} & \checkmark & - & 91.5  / 86.7 \\
     ~ & -  & \checkmark  &92.1 / 87.2  \\
      ~ & \checkmark  & \checkmark  & 91.7 / 86.8 \\
    \hline  

  \multirow{3}*{GAT} & \checkmark & -  & 92.3 / 87.4 \\
     ~ & -  & \checkmark  &  92.4 / 87.3 \\
      ~ & \checkmark  & \checkmark  & \textbf{92.7} / \textbf{87.8}  \\
    \hline  
    \end{tabular}
    }
    \caption{The influence of generating middle features from different modality and range features on the RegDB dataset.}
    \label{T-MR}
    \end{center}
    
    \end{table}

      \begin{table}[!t]
      \small
  \begin{center}

    {
    \begin{tabular}{c |c | c }
    \hline  
    \multirow{2}*{Settings}&\multicolumn{1}{c|}{SYSU-MM01}&\multicolumn{1}{c}{RegDB}\\
    \cline{2-3}
    ~ &  R-1 / mAP & R-1 / mAP\\
    \hline  
      Baseline  & 69.9 / 66.9 & 85.0 / 79.1 \\
      plugged after stage 1  & 70.6 / 66.8 & 92.5 / 87.4  \\
      plugged after stage 2   & 71.8 / 68.1 &93.4 / 88.0 \\
      plugged after stage 3   & 71.2 / 68.9 & 93.5 / 88.9 \\
      plugged after stage 4   & \textbf{75.6} / \textbf{74.2} & \textbf{94.7} / \textbf{90.4}\\
      plugged after stage 5  & 72.6 / 71.0 & 92.8 / 88.2\\

    \hline 
    \end{tabular}
    }
    \caption{The influence of the stage of backbone to plug the SLE, HSL, and CFL.}
        \label{T-Stage}

    \end{center}
    \end{table}

\begin{table*}[!ht]
\begin{center}
\small
    {
    \begin{tabular}{l c  c | c c c  }
    \hline  
    \multirow{2}*{Settings} &\multirow{2}*{FLOPs (G)}&\multirow{2}*{Params (M)} & \multicolumn{3}{c}{SYSU-MM01}\\
    \cline{4-6}
    ~ & & &~Training time (H) &Inference time (S)&mAP\\
    \hline  
    Baseline     &5.2   &23.5 & 5.4 &65& 66.9  \\
    +SLE    &7.1 &38.8 &15.7 &79&69.4   \\
    +SLE+HSL  &9.2 &51.9 &19.1 &87&72.4   \\
    +SLE+HSL+CFL   &14.3 &83.4 &32.0 &87& 72.9    \\
    +SLE+HSL+CFL+$\mathcal{L}_{MRIC}$   &14.3 &83.4 &32.4&87 & 74.2   \\
    \hline 
    \end{tabular}
}
\caption{FLOPs,  the number of parameters, training time, inference time, and mAP of the key components of our method on the SYSU-MM01 dataset.}
    \label{T4-1}

\end{center}
\end{table*}

\begin{table*}[!ht]
\begin{center}
\small
    {
    \begin{tabular}{l c  c | c c  c  }
    \hline  
    \multirow{2}*{Settings} &\multirow{2}*{FLOPs (G)}&\multirow{2}*{Params (M)} & \multicolumn{3}{c}{RegDB}\\
    \cline{4-6}
    ~ & & &~Training time (H) & Inference time (S)&mAP\\
    \hline  
    Baseline     &5.2   &23.5 & 6.1 &270& 79.1    \\
    +SLE    &7.1 &38.8 &15.3 &360& 84.8  \\
    +SLE+HSL  &9.2 &51.9  &19.7 &380& 87.1  \\
    +SLE+HSL+CFL   &14.3 &83.4  &29.0 &380& 87.8    \\
    +SLE+HSL+CFL+$\mathcal{L}_{MRIC}$   &14.3 &83.4 &31.3 &380 & 90.4    \\
    \hline 
    \end{tabular}
}

\caption{FLOPs,  the number of parameters, training time, inference time, and mAP of the key components of our method on the  RegDB dataset.}
    \label{T4-2}
\end{center}

\end{table*}

{\noindent \textbf{Influence of the Numbers of Convolutional Blocks and Transformer Blocks.} From Figure \ref{Number}, we can observe that when we adopt 3 convolutional blocks and 2 Transformer blocks in the SLE module, our method can achieve the best results.}

\noindent \textbf{Influence of the Hyperparameters.} {We evaluate the influence of the number of hyperedges $M$ and the hyperparameter $\lambda$ in Eq.~(6), as shown in Figure \ref{Lambda}.  We can observe that the best performance is achieved when the values of $M$ and $\lambda$ are set to 256 and 1.3, respectively.}

\noindent \textbf{Influence of Different Middle Features.} {We also evaluate the influence of different middle features on the RegDB dataset, and the results are shown in Table \ref{T-MR}. To obtain reliable middle features, we propose the CFL module to align different modalities and ranges features by leveraging graph attention (GAT). Our method can achieve the best mAP performance on the RegDB dataset.}

\noindent \textbf{Influence of the Stage of the Backbone to Plug the Key Components.}
Our key components (SLE, HSL, and CFL) can be easily plugged into any stage of the backbone network. In this paper, we use AGW as the backbone, which consists of five stages (stages 1-5). We plug these key components into different stages of the backbone and show the performance. 
The results are given in Table  \ref{T-Stage}. The best performance is achieved when these key components are inserted after stage 4. This indicates that the high-level semantic features extracted from stage 4 facilitate the model to explore the high-order structure information in different ranges of features and generate reliable middle features.

\noindent {\textbf{Computational Complexity of Key Components.}
We report the FLOPs, the number of parameters,  training time, inference time, and mAP of key components of our method on the SYSU-MM01 and RegDB datasets, as shown in Tables \ref{T4-1} and \ref{T4-2}, respectively.  Compared with the Baseline, the computational complexity and the number of parameters obtained by our method increase. But the performance improvements are more significant (e.g., our HOS-Net surpasses the baseline by about 7.3\% and 11.3\% in mAP for SYSU-MM01 and RegDB datasets, respectively). 

Note that the CFL module generates middle features based on the ground-truth labels of VIS and IR images. During the testing stage, the label information is not available and thus the CFL module is not used for cross-modality matching. Therefore, the inference time obtained by the methods with and without the CFL module is the same. However, by incorporating the CFL module, our method can learn a reasonable common feature space. Hence, the method with the CFL module can achieve higher performance than that without the CFL module.

\section{Supplement  D.  More Visualization Analysis}

\begin{figure}[!t]
\centering
\includegraphics[width=1.0 \columnwidth]{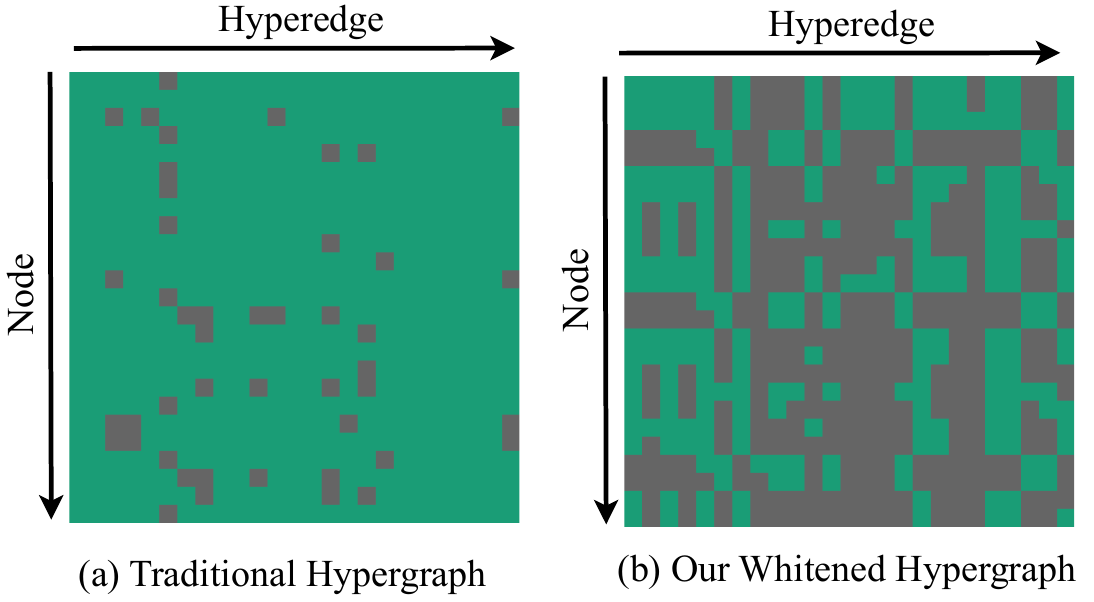}
\caption{Visualization of the high-order relationship obtained by different methods. In each column, the green square represents that the node is connected by a hyperedge, while the gray square represents that the node is without connections.}
\label{Hyper_Index}
\end{figure}

\begin{figure}[!t]
\centering
\includegraphics[width=0.8 \columnwidth]{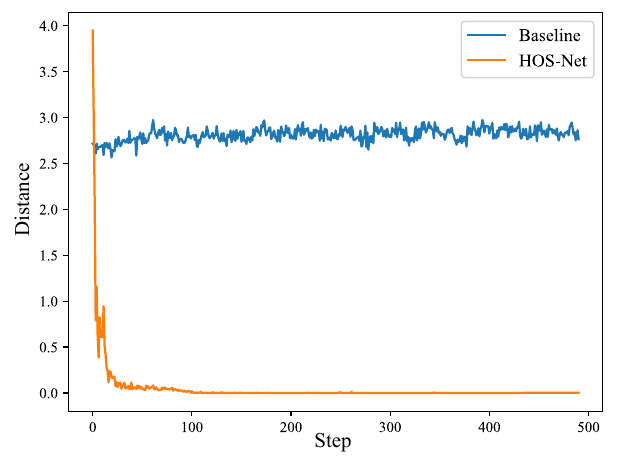}
\caption{{The distance curves between VIS and IR features when training with Baseline and HOS-Net.}}
\label{Smooth}
\end{figure}

\begin{figure}[!t]
\centering
\includegraphics[width=1.0\columnwidth]{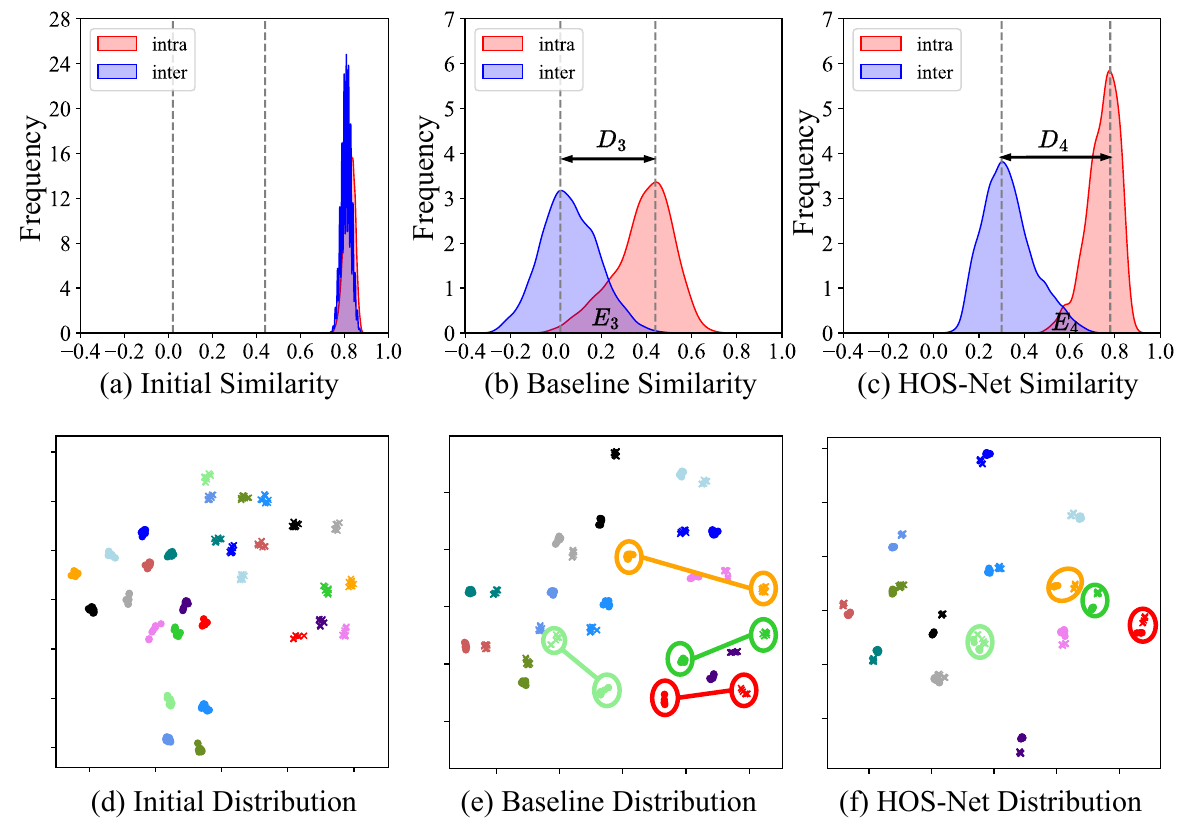}
\caption{{(a-c) demonstrate the distributions of intra-class and inter-class similarities of VIS and IR modality features on the RegDB dataset. 
(d-f) visualize the distribution of person features from VIS and IR modalities in the 2D feature space on the RegDB dataset. Circles and crosses in distinct colors represent features from VIS and IR modalities, respectively.}}
\label{Frequency_Distribution_RegDB}
\end{figure}

\begin{figure}[t]
\centering
\includegraphics[width=1.0\columnwidth]{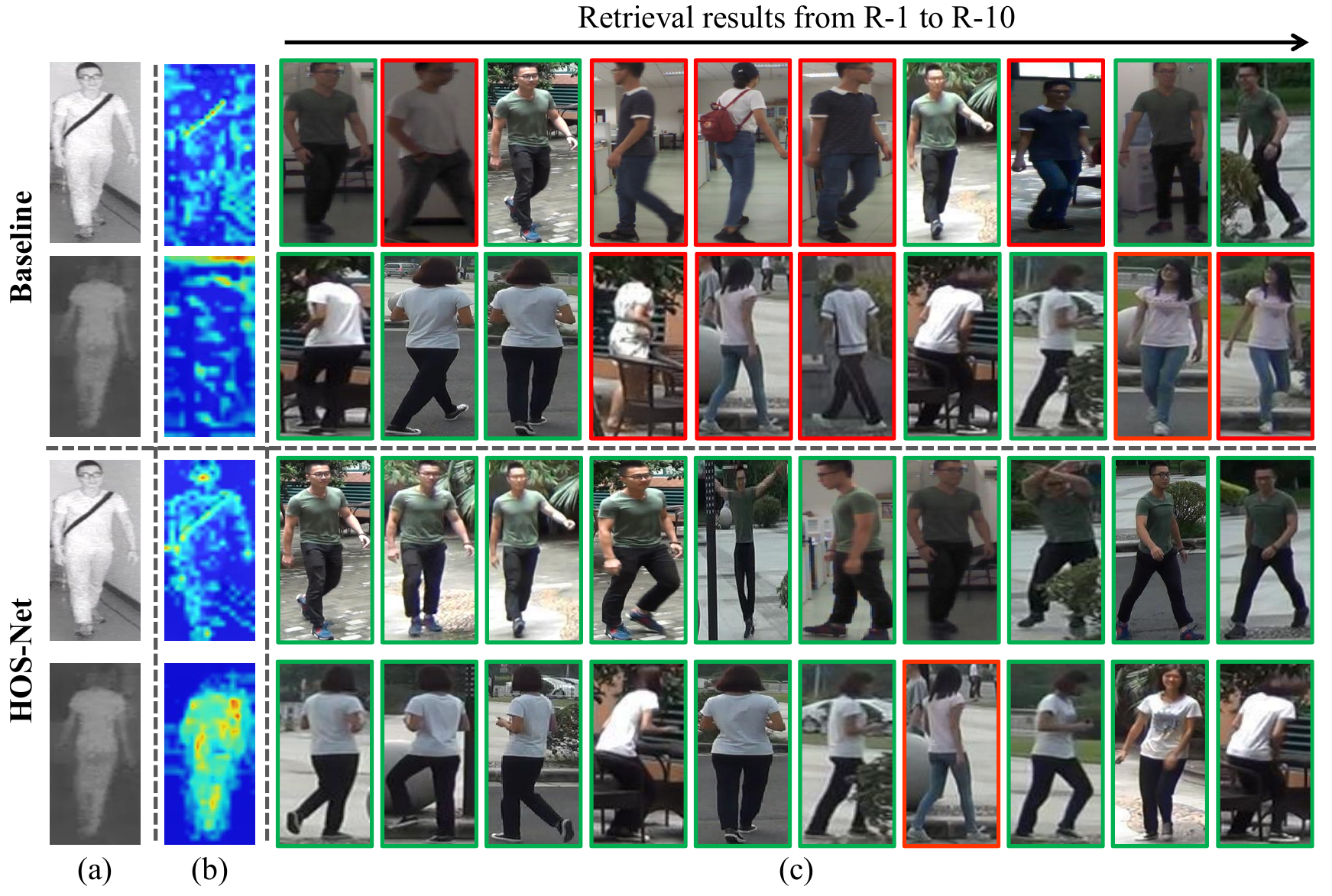}
\caption{Attention maps and retrieval results obtained by the baseline and the proposed HOS-Net on the SYSU-MM01 dataset. (a) Query images. (b) Attention maps. (c) Retrieval results {(Green: correct retrieval, Red: incorrect retrieval).}}

\label{Retrival}
\end{figure}

{\noindent \textbf{Visualization of the High-Order Relationship.} We present the visualization of the high-order relationship obtained by the traditional hypergraph network and our whitened hypergraph network, as shown in Figure \ref{Hyper_Index}.
We can see that  
many nodes share the same hyperedges for the traditional hypergraph network. Thus, diverse high-order connections are collapsed into a single connection.
Compared with the traditional hypergraph network, our whitened hypergraph network can avoid model collapse by introducing a whitening operation that projects the nodes into a spherical distribution.}

\noindent {\noindent \textbf{Distance Curve between VIS and IR Features during Training.} From Figure \ref{Smooth}, we can observe that compared with the Baseline, the proposed HOS-Net can effectively reduce the distances between the VIS and IR, greatly smoothing the process of learning the common feature space by exploiting high-order structure information and reliable middle features.

\noindent \textbf{Feature Distribution Visualization.} From Figure \ref{Frequency_Distribution_RegDB}(a-c), we can observe the cosine similarities of positive (negative) matching pairs achieved by HOS-Net are higher (lower) than the Baseline on the RegDB dataset. Compared with the Baseline, HOS-Net can also reduce the intra-class differences and enlarge inter-class gaps in different modalities more effectively, as shown in Figure \ref{Frequency_Distribution_RegDB}(d-e).

\noindent \textbf{Retrieval Results.} To further evaluate the effectiveness of our proposed HOS-Net, we show the attention maps and retrieval results obtained by our method and the baseline on the SYSU-MM01 dataset. As shown in Figure \ref{Retrival}(b), unlike the Baseline, our HOS-Net shows the superior ability to focus on discriminative pedestrian features, mitigating the influence caused by pose variations, background interference, and modality discrepancy. Figure \ref{Retrival}(c) shows that our HOS-Net achieves more precise retrieval results in different modalities than the baseline.

\end{document}